\documentclass[runningheads]{llncs}
\usepackage[T1]{fontenc}
\usepackage{xcolor}
\usepackage{graphicx,verbatim}
\usepackage{booktabs} 
\usepackage{multirow} 
\usepackage{hyperref} 
\hypersetup{
    colorlinks=true,
    citecolor=blue
}
\begin{document}
\title{RetinalGPT: A Retinal Clinical Preference Conversational Assistant Powered by Large Vision-Language Models}
%

\def\thefootnote{*}\footnotetext{These authors contributed equally to this paper.}
\def\thefootnote{$\dagger$}\footnotetext{Corresponding author: wzhu59@asu.edu}
\author{Wenhui Zhu\inst{1* \dagger} \and
Xin Li\inst{1*} \and Xiwen Chen\inst{2}  \and
Peijie Qiu\inst{3}  \and  Vamsi Krishna Vasa\inst{1} \and Xuanzhao Dong\inst{1} \and
Yanxi Chen \inst{1} \and Natasha Lepore\inst{4}
Oana Dumitrascu \inst{5} \and Yi Su\inst{6} \and Yalin Wang\inst{1}
}
\authorrunning{Zhu. Wenhui et al.}
%
\institute{
School of Computing and Augmented Intelligence, Arizona State University, AZ, USA \and
School of Computing, Clemson University, SC, USA \and
Washington University School of Medicine in St. Louis, MO, USA \and 
The Keck School of Medicine of the University of Southern California, CA, USA \and
Mayo Clinic, AZ, USA \and
Banner Alzheimer’s Institute, AZ, USA\\
}

\maketitle              
\begin{abstract}
Recently, Multimodal Large Language Models (MLLMs) have gained significant attention for their remarkable ability to process and analyze non-textual data, such as images, videos, and audio. Notably, several adaptations of general-domain MLLMs to the medical field have been explored, including LLaVA-Med. However, these medical adaptations remain insufficiently advanced in understanding and interpreting retinal images. In contrast, medical experts emphasize the importance of quantitative analyses for disease detection and interpretation. This underscores a gap between general-domain and medical-domain MLLMs: while general-domain MLLMs excel in broad applications, they lack the specialized knowledge necessary for precise diagnostic and interpretative tasks in the medical field. 
To address these challenges, we introduce \textit{RetinalGPT}, a multimodal conversational assistant for clinically preferred quantitative analysis of retinal images. Specifically, we achieve this by compiling a large retinal image dataset, developing a novel data pipeline, and employing customized visual instruction tuning to enhance both retinal analysis and enrich medical knowledge. In particular, RetinalGPT outperforms MLLM in the generic domain by a large margin in the diagnosis of retinal diseases in 8 benchmark retinal datasets.
Beyond disease diagnosis, RetinalGPT features quantitative analyses and lesion localization, representing a pioneering step in leveraging LLMs for an interpretable and end-to-end clinical research framework.The code is available at \url{https://github.com/Retinal-Research/RetinalGPT}



\end{abstract}
\section{Introduction}

Retinal color fundus photography (CFP) plays a crucial role in diagnosing ocular diseases, with non-mydriatic CFP becoming increasingly popular for point-of-care diagnostics~\cite{wolf2020cost}. Over the past decades, deep neural networks, especially convolutional neural networks (CNNs), have been widely applied in retinal image analysis, delivering state-of-the-art performance across various retinal disease (RD) detection tasks~\cite{zhu2025eyebenchrigorousevaluationretinal,zhu2024nnmobilenetrethinkingcnnretinopathy,vasa2024contextawareoptimaltransportlearning,dong2024cunsbrfiecontextawareunpairedneural}. As the field advances, attention is increasingly shifting toward combining visual and textual data to unlock new possibilities~\cite{grattafiori2024llama3herdmodels,liu2023visualinstructiontuning,li2023llavamedtraininglargelanguageandvision,openai2024gpt4technicalreport}.

Vision-language models (VLMs) integrate visual and textual data, enabling computers to process and analyze multimodal information~\cite{zhang2024generalist}. The widespread availability of image-text pairs has facilitated self-supervised training, as demonstrated by multimodal GPT-4~\cite{openai2024gpt4technicalreport} and open-source models like LLaVA~\cite{liu2023visualinstructiontuning}. By instruction-tuning these models to align with human preference through multimodal inputs they achieve strong zero-shot performance in tasks such as image interpretation and reasoning, paving the way for versatile multimodal conversational assistants~\cite{askell2021generallanguageassistantlaboratory,li2022elevaterbenchmarktoolkitevaluating,gan2022visionlanguagepretrainingbasicsrecent}.
However, their reliance on generic-domain knowledge often restricts their effectiveness in specialized tasks, such as medical diagnosis, where domain-specific expertise is crucial. To overcome this limitation, LLaVA-Med~\cite{li2023llavamedtraininglargelanguageandvision} employs a multimodal instruction-tuning strategy, enabling comprehensive end-to-end training of a specialized biomedical conversational assistant. By leveraging the advanced capabilities of GPT-4, it generates a diverse set of biomedical instruction-following data from image-text pairs within PMC-15M~\cite{zhang2025biomedclipmultimodalbiomedicalfoundation}. Additionally, it fine-tunes a vision-language model specifically for biomedical applications using a novel curriculum learning approach, ensuring improved understanding and accuracy in medical contexts.

However, existing medical-domain VLMs still suffer from certain limitations when it comes to handling specialized downstream tasks, such as clinical analysis in retinal images. Medical experts often employ a \textit{morphological} method to analyze internal feature changes and potential biomarker discovering, i.e. the impact of vascular branch features in retinal images on Alzheimer's disease~\cite{ad,ad1}.
In contrast, existing medical-domain VLMs struggle with clinical preference reasoning, as they are not trained on such structured analytical instruction data. This also leads to their inability to effectively localize accurate abnormalities such as lesions. Another significant limitation of current medical-domain VLMs is the inefficiency of instruction-tuning techniques, which often leads to a loss of generic-domain knowledge.

\begin{figure}[htp]
  \centering
  \centerline{\includegraphics[width=0.85\textwidth]{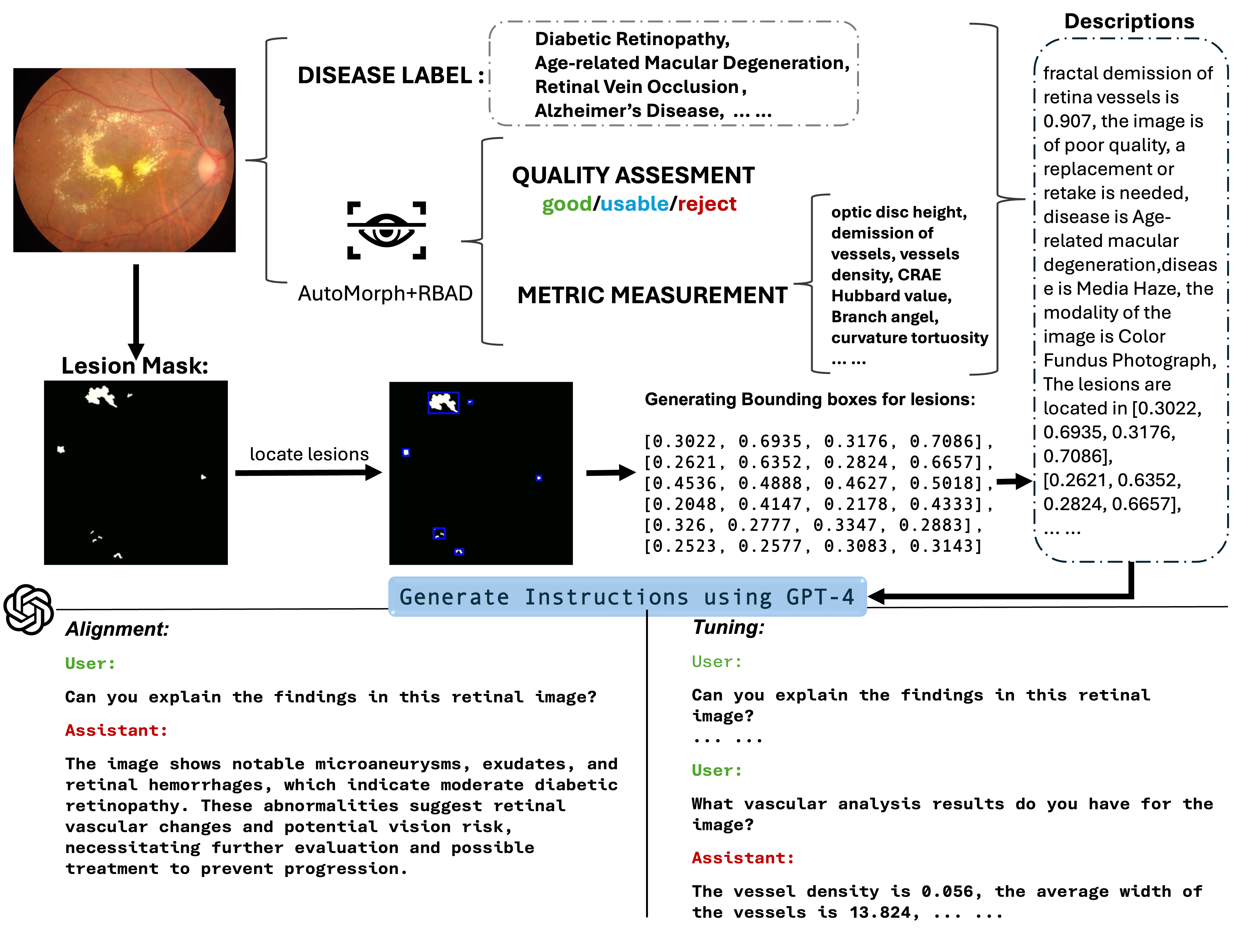}}
\caption{\textbf{Top:} the process of obtaining data. \textbf{Bottom:} The collected data is converted into instruction-following data, categorized into Alignment (\textbf{Bottom Left}) and Tuning data (\textbf{Bottom Right}).
}
\label{fig:data_collect}
\end{figure}

To overcome these limitations, we introduce an end-to-end pipeline that focuses on constructing an efficient clinical dataset for retinal image analysis and enhancing the instruction-tuning technique. The proposed instruction-tuning technique leverages two-stage approach for adapting general-domain VLMs for retinal domain tasks. First, we broaden the vocabulary of aligned image-text tokens to incorporate both retinal-specific knowledge and generic biomedical information, which preserves the broader knowledge. Second, we perform \textbf{Mixup} instruction-tuning on domain-specific knowledge and generic knowledge, which retains the broader knowledge space. The major contributions of this paper are:
\textbf{(i)} To the best of our knowledge, we are the first to advance VLMs for specialized downstream tasks in retinal image analysis by curating clinical preference data to enable more effective instruction tuning.
\textbf{(ii)} Our proposed method not only enhances the knowledge domain of the models to be leveraged for specific downstream tasks, but also preserves generic medical domain knowledge (refer Section \ref{exp}). 
\textbf{(iii)} RetinalGPT demonstrated superior performance across multiple downstream tasks on eight publicly available retinal image datasets.


\section{Method}

\subsection{Visual Instruction-Following Data}

\noindent\textbf{Clinical Data Extraction.}
Here, We collected retinal images (open access and private data) for extracting clinical feature descriptions, which include disease labels, lesion bounding boxes and fractal features. The clinical data extraction is illustrated in Fig.~\ref{fig:data_collect}. Specifically, we collected the disease label for each image based on the medical annotations. Each image was then processed using AutoMorph~\cite{automorph} and RBAD~\cite{rbad}, which performed fractal analysis of the vascular structures. This analysis extracted features such as fractal dimension of vessels, vessel density, branch angle, curvature tortuosity, as well as other morphological characteristics relevant to vascular structure. To ensure the extracted features were clinically relevant, we used statistical analysis~\cite{retinalfractal}, to examine their relationship with disease labels. We also consulted ophthalmologists to confirm the importance of each feature, keeping those that significantly contributed to disease characterization while removing redundant or less relevant ones. Subsequently, each dataset retained around 40 important features for further analysis. Additionally, AutoMorph~\cite{automorph} also evaluated image quality $X_{q}$ , assigning one of three labels, good, usable and reject. For datasets containing lesion masks, such as IDRiD~\cite{idrid} and MICCAI MMAC~\cite{miccai2023mmac}, we located lesions and generated bounding boxes in the format $(x_{min},y_{min},x_{max},y_{max})$. Finally, we integrated the extracted disease labels ${X_d}$, quality assessments $X_{q}$, vascular features ${X_f}$, lesion information ${X_l}$ and the modality of the image $X_m$(Color Fundus Photograph) into a structured description for each image, as ($X_{f},X_{q},X_{d},X_{m},X_{l}$). After filtering out low-quality images that could not generate valid analysis reports, we prepared approximately 38K images along with their corresponding structured descriptions to construct our data in the following.

\noindent\textbf{Retinal Concepts Alignment Data.}
For each image, we generate single-round instruction-following dialogues about the image using the disease label ${X_d}$ and the image modality ${X_m}$, without utilizing the image, by leveraging GPT-4~\cite{openai2024gpt4technicalreport}, shown at the bottom of Fig.~\ref{fig:data_collect}. We constructed a sample question ${X_q}$ and got the answer ${X_m,X_d}$ as follows:

\begin{center}
\textbf{Human}: \(X_q\) <STOP>

\textbf{Assistant}: \(X_m X_d\) <STOP>
\end{center}

The generated question ${X_q}$ is designed to provide a concise and generic inquiry about information of the image. We generated 38K question-answer pairs as the Retinal Concepts Alignment(38K RCA) Dataset. To enhance the generic medical understanding, we combined 600K image-text pairs from PMC-15M (600K PMC)~\cite{PMC-15}, allowing the model to learn a wider range of medical expressions and improve its ability to generate medically precise descriptions.

\noindent\textbf{Clinical Preference Instruction-Tuning Data.}
To align the model with biomedical instruction-following tasks, we generated multi-turn conversations using GPT-4~\cite{openai2024gpt4technicalreport} following the same format as the alignment process, prompting it to simulate responses as if it had access to the image, even though it only utilized text information. At this stage, we used all extracted image information to generate instruction data, ($X_{f},X_{q},X_{d},X_{m},X_{l}$), examples shown at the bottom of the Fig.~\ref{fig:data_collect}. To ensure high-quality instruction data, we filtered out cases where fractal-based vascular features are insufficient, as they may not provide meaningful context for question generation. Additionally, we designed specific prompts to guide GPT-4~\cite{openai2024gpt4technicalreport} in generating targeted question-answer pairs tailored to different datasets. These prompts are customized to adapt to various tasks, such as DR grading, multi-disease classification, and lesion localization, ensuring the generated instructions align with the characteristics of each dataset. As a result, we generated multi-turn instruction-following dialogues for 38K retina fundus images as the Retinal Concepts Tuning (38K RCT) Dataset. To improve the ability to generalize across generic medical domains, we integrated 60K image-text QA pairs (60K Generic QA) from five common medical imaging modalities, including CXR, CT, MRI, histopathology, and gross pathology, from LLAVA-Med~\cite{li2023llavamedtraininglargelanguageandvision}.

\begin{figure}[!h]
  \centering
  \centerline{\includegraphics[width=\textwidth]{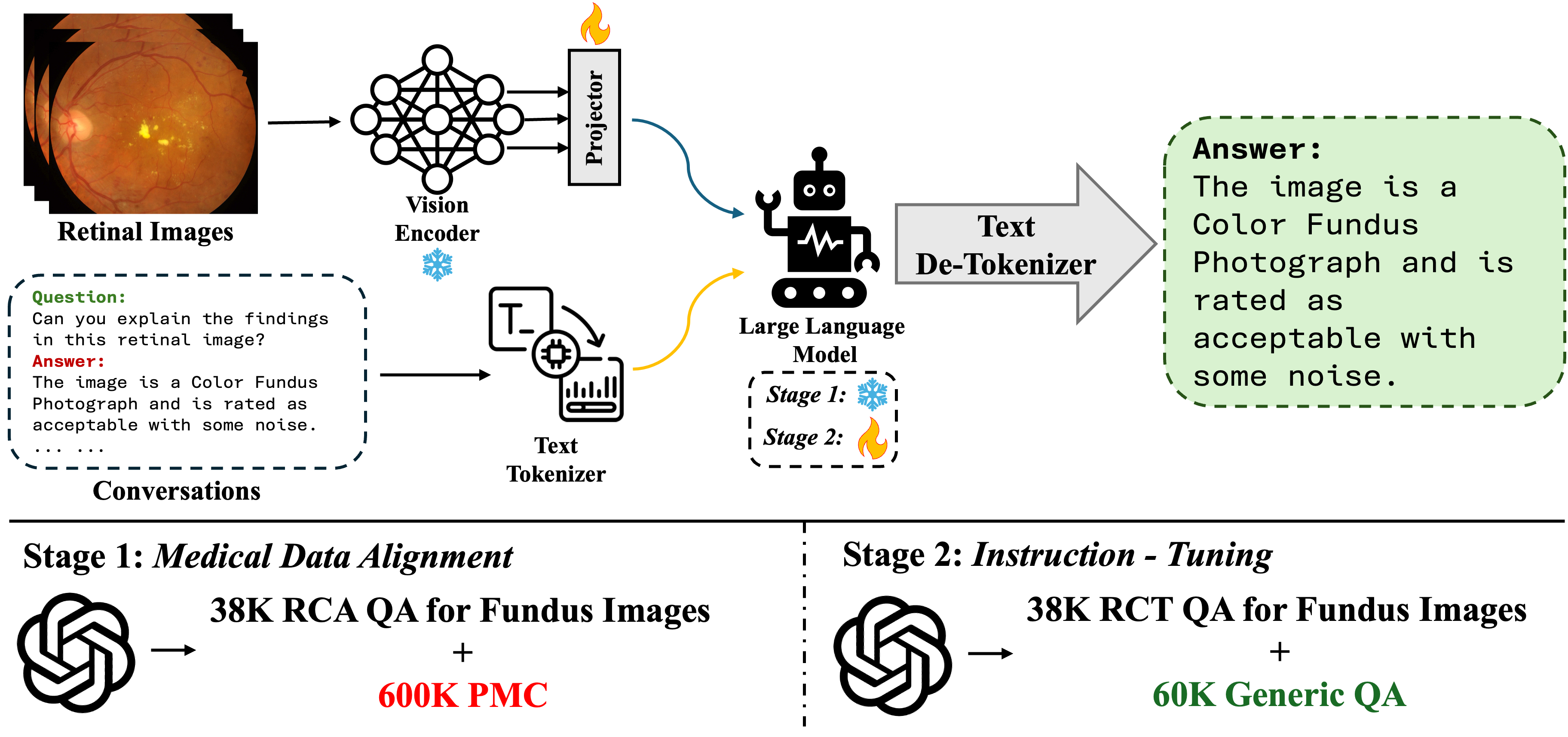}}
\caption{Overview of the model architecture and training strategy. \textbf{Top:} The network structure. \textbf{Bottom:} the two-stage training process.
}
\label{fig:arch}
\end{figure}

\subsection{Training Strategies} We adopted the architecture employed by LLaVA and LLaVA-Med for our MLLM where a linear projection layer connects the vision encoder and the language model, and initialied the model parameters using pretrained weights. Here, our objective is to preserve the knowledge of the generic biomedical domain, while adapting the retinal-specific domain. To achieve this, we implement a two-stage training strategy utilizing mixup dataset, as shown in Fig~\ref{fig:arch}. 

\noindent\textbf{Stage 1 - Feature Alignment.} Building upon the method of LLaVA, we align image-text pairs by leveraging a mixup of the PMC-600K and 38K RCA aligment datasets to retain generic medical domain knowledge. During training, we freeze the weights of both the vision model (CLIP) and the LLM (LLama), updating only the projection matrix. Our aim is to align the image features of retinal images and other biomedical visual concepts with their corresponding textual word embeddings from the pre-trained language model. This process expands the vocabulary of aligned image-text tokens to include both retinal-specific knowledge and broader biomedical information.

\noindent\textbf{Stage 2 - Mixup Instruction-Tuning.} In this stage, we frozen the vision model, the projector layer and LLM only are only trained, and both model initial pretrain weight from stage-1.  To enable RetinalGPT to understand and process retinal specific domain knowledge, it is trained with the 38k RCT QA clinical preference instruction-instruction data generated in Section 2.1. However, solely relying on our 38K retinal domain data is insufficient for the language model to acquire broader medical knowledge and may result in hallucinations, thereby causing the model to overfit to the retinal medical domain. To enhance RetinalGPT's capabilities in visual understanding and question answering, we expand the fine-tuning dataset by incorporating 60K Generic QA generated by LLaVA-Med. In this context, \textit{Mixup} indicates that the training process integrates both generic medical conception understanding data and task-specific instruction tuning retinal data. Consequently, RetinalGPT is finetuned with 98K image-text instruction-following data for clinical retinal together with 60K generic medical instruction-following data. The former ensures that RetinalGPT can be applied for clinical Preference retinal knowledge, while the latter enhances the medical data diversity and general visual understanding ability of RetinalGPT.

\noindent\textbf{Implementation Details.} Both two stages training on 4 A100 80G GPUs with a global batch size of 128, max length 2048, and 0 weight decay, where Stage 1 runs for 10 epoch at a learning rate of $2e^{-3}$, and Stage 2 runs for 10 epochs at a learning rate of $2e^{-5}$.

\section{Experiment}\label{exp}
\subsection{Evaluation Datasets}
In order to enhance the generalization of our model, we collected multiple public color fundus image datasets to ensure data diversity and representativeness. We utilized 3,772 images from various datasets, covering a wide range of ophthalmic diseases. Specifically, \textbf{Messidor-1}~\cite{messidor2014messidor}, \textbf{APTOS}~\cite{aptos2019}, and \textbf{EyeQ}~\cite{eyeq} for diabetic retinopathy (DR) grading, with EyeQ further providing image quality labels. \textbf{IDRiD}~\cite{idrid} and \textbf{MICCAI MACC}~\cite{miccai2023mmac} for DR lesion and myopic maculopathy analysis. \textbf{OIA-ODIR}~\cite{odir} and \textbf{RFMiD}~\cite{rfmid} cover various ophthalmic diseases, such as glaucoma, cataract, hypertensive retinopathy, and others. Additionally, we included a \textbf{Private} dataset, which include cases of Alzheimer’s disease (AD). These datasets were never used to train our model, ensuring that the evaluation set remains completely independent and provides an unbiased assessment of the model’s performance.

\subsection{Multi-Disease Abnormal Detection}
To evaluate the effectiveness of abnormality detection, we selected state-of-the-art large multimodal language models and performed abnormality classification on our evaluation datasets by uploading images and querying whether abnormalities were present. The results are presented in Table.~\ref{tab:baseline_comparison}. As shown in the table, our model achieves the best performance across all evaluation datasets, regardless of whether the task involves multi-disease classification or single-disease DR classification. Despite variations in image quality across different datasets, our method consistently outperforms all baselines, demonstrating its robustness in fundus abnormality classification.

\begin{figure}[!h]
  \centering
  \centerline{\includegraphics[width=0.75\textwidth]{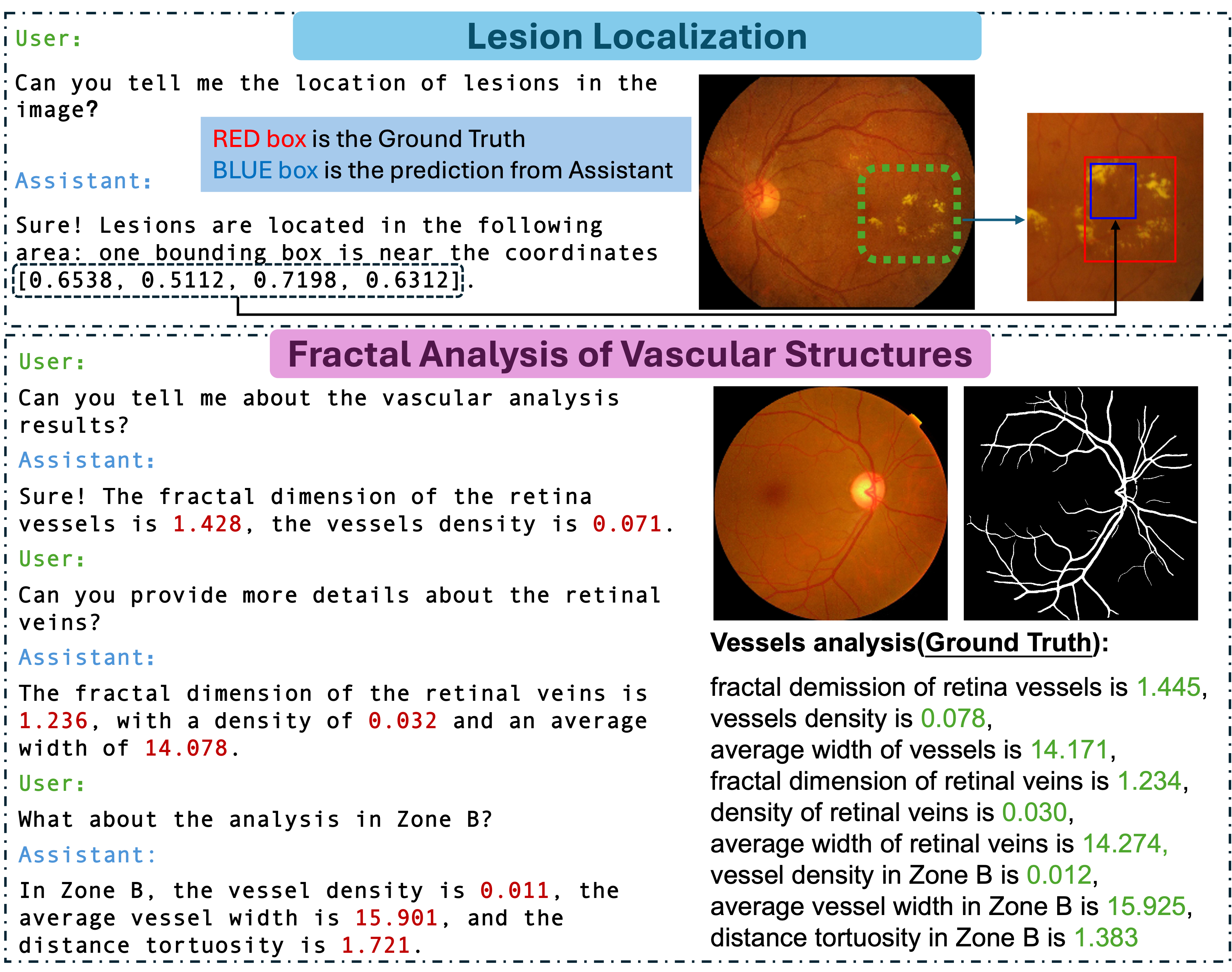}}
\caption{The top section compares the our model can predict lesion locations (blue box) with ground truth annotations (red box). The bottom section presents vascular structure analysis, where the model can estimate fractal dimension, vessel density, and vessel width are compared against ground truth values.}
\label{fig:localize&fractal}
\end{figure}

\begin{table}[h]
    \centering
    \renewcommand{\arraystretch}{1.3} 
    \resizebox{\textwidth}{!}{ 
    \begin{tabular}{c|cccccccc}  
        \toprule
        \textbf{Method} & \textbf{APTOS } & \textbf{EyePACS } & \textbf{IDRID } & \textbf{Messidor } & \textbf{MICCAI } & \textbf{OIA-ODIR } & \textbf{RFMiD } & \textbf{Private} \\
        \midrule
        Llama-3.2-11B-Vision  & 59.78 & 60.27 & 47.17 & 59.17 & 42.86 & 44.62 & 31.27 & 13.85 \\ 
        LLaVA                 & 56.52 & 43.13 & 56.6 & 54.17 & 56.43 & 55.38 & 65.15 & 55.41  \\ 
        LLaVA-med             & 47.00 & 55.57 & 49.01 & 53.33 & 57.14 & 48.92 & 53.09 & 44.16    \\
        GPT-4o                & 54.35 & 53.03 & 52.83 & 58.33 & 45.71 & 52.69 & 58.63 & 31.17 \\ \hline
        Ours                  & \textcolor{red}{95.10} & \textcolor{red}{73.70} & \textcolor{red}{84.90} & \textcolor{red}{80.83} & \textcolor{red}{87.14} & \textcolor{red}{66.13} & \textcolor{red}{88.27} & \textcolor{red}{99.57}   \\ 
        \bottomrule
    \end{tabular}
    }
    \caption{Performance comparison of different models on abnormality classification.}
    \label{tab:baseline_comparison}
\end{table}

\subsection{Clinical Preference Conversation Interaction}
Our model is capable of performing both lesion localization and vascular structure analysis by following instruction-based prompts, allowing it to provide interpretable and structured outputs. 

\noindent \textbf{Lesion Localization}
When prompted about lesion locations, our model is able to identify their approximate positions, such as near the macula. Furthermore, the model provides precise bounding box coordinates for lesion localization. As shown in the top section of Fig.~\ref{fig:localize&fractal}. We evaluated this capability by querying lesion locations in images with disease masks and comparing the predicted bounding boxes (blue) with the ground truth lesion masks (red). The results show that the predicted locations closely align with actual lesion regions.

\noindent \textbf{Vascular Structure Analysis}
Our model performs vascular structure analysis, returning key feature values when queried. It can also provide detailed vascular metrics for specific regions upon further inquiry. As shown in the bottom section of Fig.~\ref{fig:localize&fractal}, we compared the predicted values of the model with those computed by AutoMorph~\cite{automorph}, and the results show that the values are very close, demonstrating the accuracy of the model in vascular feature estimation.

\begin{figure}[h]
  \centering
\centerline{\includegraphics[width=0.87\textwidth]{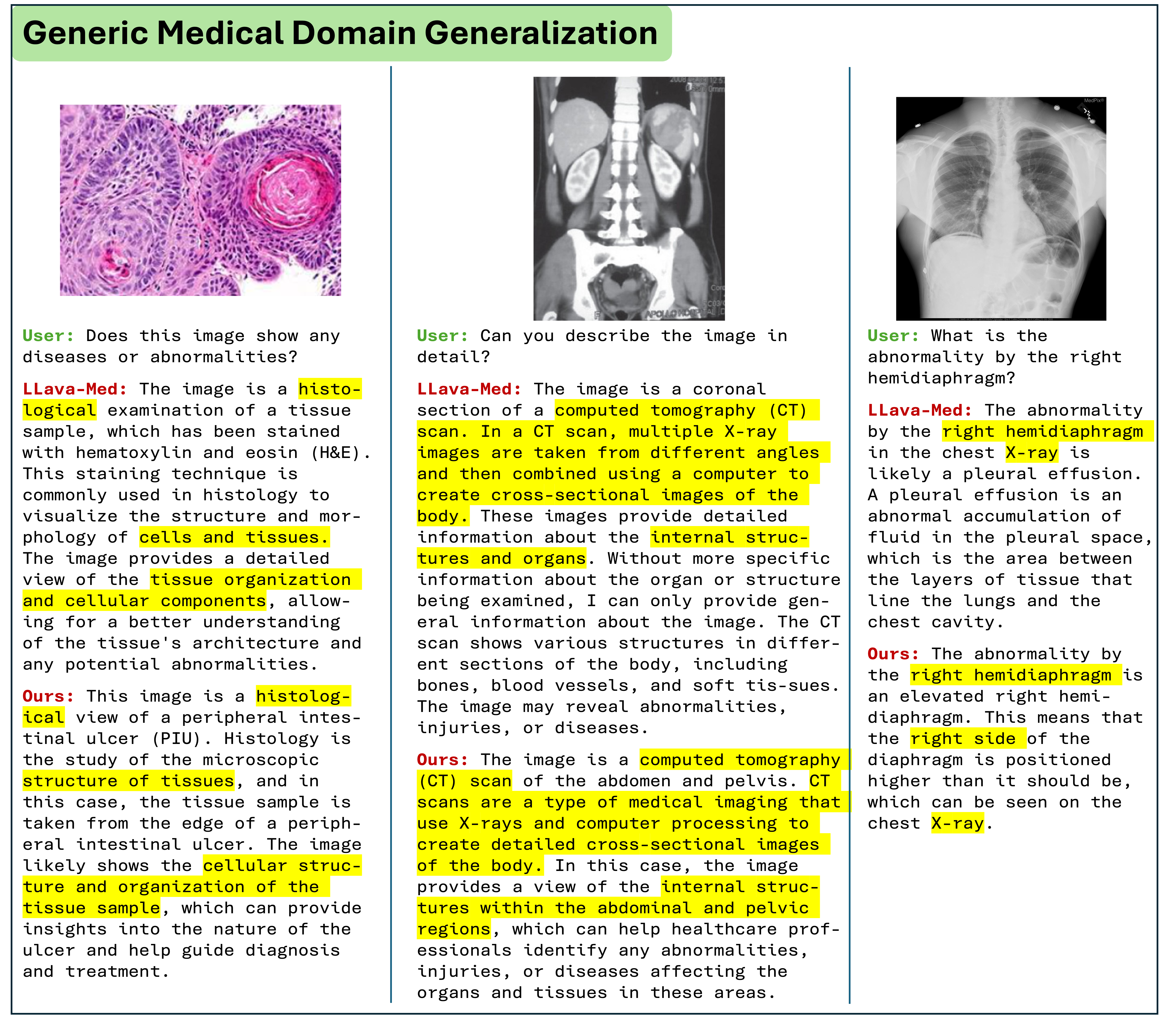}}
\caption{Comparing our model and LLaVA-Med~\cite{li2023llavamedtraininglargelanguageandvision} on generic medical domain. The highlighted sections illustrate highly similar parts of the responses. Our model demonstrates ability to preserve knowledge in the generic medical domain.}
\label{fig:overview}
\end{figure}

\subsection{Generic Medical Domain Generalization}
Our model demonstrates strong performance not only in retinal fundus imaging but also in generic medical domain. To assess its performance in this task, we conducted evaluations on generic medical questions from LLaVA-Med~\cite{li2023llavamedtraininglargelanguageandvision} using both our model and LLaVA-Med. As shown in Figure 4, our model and LLaVA-Med~\cite{li2023llavamedtraininglargelanguageandvision} produce highly similar responses when tested on different medical imaging modalities, including histological slides, CT scans, and X-rays.

\section{Conclusion}
In this paper, we introduce RetinalGPT, a MLLM designed specifically for analyzing retinal images. It uses a large dataset, a new data pipeline, and custom visual instructions to improve retinal disease diagnosis. RetinalGPT not only diagnoses diseases accurately but also provides detailed analysis and locates lesions. Notably, it also holds generic medical knowledge. By combining different data types into a clear clinical framework, it helps advance research and the use of large language models in medical imaging. During our evaluation, we found that the assistant tends to give modality-related answers for the first question in retinal images, no matter what is asked, while answering later questions normally. This may be due to a lack of diversity in retinal QA instruction tuning set. We plan to focus on solving this issue in our future work.

\bibliographystyle{splncs04}
\bibliography{ref}

\begin{thebibliography}{10}
\providecommand{\url}[1]{\texttt{#1}}
\providecommand{\urlprefix}{URL }
\providecommand{\doi}[1]{https://doi.org/#1}

\bibitem{miccai2023mmac}
{MICCAI MMAC 2023 - Myopic Maculopathy Analysis Challenge} (2023), \url{https://codalab.lisn.upsaclay.fr/competitions/12477}

\bibitem{askell2021generallanguageassistantlaboratory}
Askell, A., Bai, Y., Chen, A., Drain, D., Ganguli, D., Henighan, T., Jones, A., Joseph, N., Mann, B., DasSarma, N., Elhage, N., Hatfield-Dodds, Z., Hernandez, D., Kernion, J., Ndousse, K., Olsson, C., Amodei, D., Brown, T., Clark, J., McCandlish, S., Olah, C., Kaplan, J.: A general language assistant as a laboratory for alignment (2021), \url{https://arxiv.org/abs/2112.00861}

\bibitem{ad1}
Cheung, C.Y., Ran, A.R., Wang, S., Chan, V.T.T., Sham, K., Hilal, S., Venketasubramanian, N., Cheng, C.Y., Sabanayagam, C., Tham, Y.C., Schmetterer, L., McKay, G.J., Williams, M.A., Wong, A., Au, L.W.C., Lu, Z., Yam, J.C., Tham, C.C., Chen, J.J., Dumitrascu, O.M., Heng, P.A., Kwok, T.C.Y., Mok, V.C.T., Milea, D., Chen, C.L.H., Wong, T.Y.: A deep learning model for detection of alzheimer's disease based on retinal photographs: a retrospective, multicentre case-control study. The Lancet Digital Health  \textbf{4}(11),  e806--e815 (Nov 2022). \doi{10.1016/S2589-7500(22)00169-8}, \url{https://doi.org/10.1016/S2589-7500(22)00169-8}

\bibitem{dong2024cunsbrfiecontextawareunpairedneural}
Dong, X., Vasa, V.K., Zhu, W., Qiu, P., Chen, X., Su, Y., Xiong, Y., Yang, Z., Chen, Y., Wang, Y.: Cunsb-rfie: Context-aware unpaired neural schr\"odinger bridge in retinal fundus image enhancement (2024), \url{https://arxiv.org/abs/2409.10966}

\bibitem{ad}
Dumitrascu, O.M., Li, X., Zhu, W., Woodruff, B.K., Nikolova, S., Sobczak, J., Youssef, A., Saxena, S., Andreev, J., Caselli, R.J., Chen, J.J., Wang, Y.: Color fundus photography and deep learning applications in alzheimer disease. Mayo Clinic Proceedings: Digital Health  \textbf{2}(4),  548--558 (2024). \doi{https://doi.org/10.1016/j.mcpdig.2024.08.005}, \url{https://www.sciencedirect.com/science/article/pii/S2949761224000804}

\bibitem{retinalfractal}
Dumitrascu, O.M., Li, X., Zhu, W., Woodruff, B.K., Nikolova, S., Sobczak, J., Youssef, A., Saxena, S., Andreev, J., Caselli, R.J., et~al.: Color fundus photography and deep learning applications in alzheimer disease. Mayo Clinic Proceedings: Digital Health  \textbf{2}(4),  548--558 (2024)

\bibitem{eyeq}
Fu, H., Wang, B., Shen, J., Cui, S., Xu, Y., Liu, J., Shao, L.: Evaluation of retinal image quality assessment networks in different color-spaces. In: Medical Image Computing and Computer Assisted Intervention--MICCAI 2019: 22nd International Conference, Shenzhen, China, October 13--17, 2019, Proceedings, Part I 22. pp. 48--56. Springer (2019)

\bibitem{gan2022visionlanguagepretrainingbasicsrecent}
Gan, Z., Li, L., Li, C., Wang, L., Liu, Z., Gao, J.: Vision-language pre-training: Basics, recent advances, and future trends (2022), \url{https://arxiv.org/abs/2210.09263}

\bibitem{grattafiori2024llama3herdmodels}
Grattafiori, A., Dubey, A., Jauhri, A., ...: The llama 3 herd of models (2024), \url{https://arxiv.org/abs/2407.21783}

\bibitem{li2022elevaterbenchmarktoolkitevaluating}
Li, C., Liu, H., Li, L.H., Zhang, P., Aneja, J., Yang, J., Jin, P., Hu, H., Liu, Z., Lee, Y.J., Gao, J.: Elevater: A benchmark and toolkit for evaluating language-augmented visual models (2022), \url{https://arxiv.org/abs/2204.08790}

\bibitem{li2023llavamedtraininglargelanguageandvision}
Li, C., Wong, C., Zhang, S., Usuyama, N., Liu, H., Yang, J., Naumann, T., Poon, H., Gao, J.: Llava-med: Training a large language-and-vision assistant for biomedicine in one day (2023), \url{https://arxiv.org/abs/2306.00890}

\bibitem{odir}
Li, N., Li, T., Hu, C., Wang, K., Kang, H.: A benchmark of ocular disease intelligent recognition: One shot for multi-disease detection. arXiv preprint arXiv:2102.07978  (2021)

\bibitem{liu2023visualinstructiontuning}
Liu, H., Li, C., Wu, Q., Lee, Y.J.: Visual instruction tuning (2023), \url{https://arxiv.org/abs/2304.08485}

\bibitem{messidor2014messidor}
MESSIDOR, T.V.: Messidor: methods to evaluate segmentation and indexing techniques in the field of retinal ophthalmology. 2014. Available on: http://messidor. crihan. fr/index-en. php Accessed: October  \textbf{9} (2014)

\bibitem{openai2024gpt4technicalreport}
OpenAI, et. al: Gpt-4 technical report (2024), \url{https://arxiv.org/abs/2303.08774}

\bibitem{rfmid}
Pachade, S., Porwal, P., Thulkar, D., Kokare, M., Deshmukh, G., Sahasrabuddhe, V., Giancardo, L., Quellec, G., M{\'e}riaudeau, F.: Retinal fundus multi-disease image dataset (rfmid): A dataset for multi-disease detection research. Data  \textbf{6}(2), ~14 (2021)

\bibitem{idrid}
Porwal, P., Pachade, S., Kamble, R., Kokare, M., Deshmukh, G., Sahasrabuddhe, V., Meriaudeau, F.: Indian diabetic retinopathy image dataset (idrid): a database for diabetic retinopathy screening research. Data  \textbf{3}(3), ~25 (2018)

\bibitem{aptos2019}
Society, A.P.T.O.: {APTOS} 2019 blindness detection dataset (2019), \url{https://www.kaggle.com/competitions/aptos2019-blindness-detection}

\bibitem{vasa2024contextawareoptimaltransportlearning}
Vasa, V.K., Qiu, P., Zhu, W., Xiong, Y., Dumitrascu, O., Wang, Y.: Context-aware optimal transport learning for retinal fundus image enhancement (2024), \url{https://arxiv.org/abs/2409.07862}

\bibitem{rbad}
Wang, H., Zhu, W., Qin, J., Li, X., Dumitrascu, O., Chen, X., Qiu, P., Razi, A.: Rbad: A dataset and benchmark for retinal vessels branching angle detection. arXiv preprint arXiv:2407.12271  (2024)

\bibitem{wolf2020cost}
Wolf, R.M., Channa, R., Abramoff, M.D., Lehmann, H.P.: Cost-effectiveness of autonomous point-of-care diabetic retinopathy screening for pediatric patients with diabetes. JAMA ophthalmology  \textbf{138}(10),  1063--1069 (2020)

\bibitem{zhang2024generalist}
Zhang, K., Zhou, R., Adhikarla, E., Yan, Z., Liu, Y., Yu, J., Liu, Z., Chen, X., Davison, B.D., Ren, H., et~al.: A generalist vision--language foundation model for diverse biomedical tasks. Nature Medicine pp. 1--13 (2024)

\bibitem{PMC-15}
Zhang, S., Xu, Y., Usuyama, N., Bagga, J., Tinn, R., Preston, S., Rao, R., Wei, M., Valluri, N., Wong, C., et~al.: Large-scale domain-specific pretraining for biomedical vision-language processing. arXiv preprint arXiv:2303.00915  \textbf{2}(3), ~6 (2023)

\bibitem{zhang2025biomedclipmultimodalbiomedicalfoundation}
Zhang, S., Xu, Y., Usuyama, N., Xu, H., Bagga, J., Tinn, R., Preston, S., Rao, R., Wei, M., Valluri, N., Wong, C., Tupini, A., Wang, Y., Mazzola, M., Shukla, S., Liden, L., Gao, J., Crabtree, A., Piening, B., Bifulco, C., Lungren, M.P., Naumann, T., Wang, S., Poon, H.: Biomedclip: a multimodal biomedical foundation model pretrained from fifteen million scientific image-text pairs (2025), \url{https://arxiv.org/abs/2303.00915}

\bibitem{automorph}
Zhou, Y., Wagner, S.K., Chia, M.A., Zhao, A., Xu, M., Struyven, R., Alexander, D.C., Keane, P.A., et~al.: Automorph: automated retinal vascular morphology quantification via a deep learning pipeline. Translational vision science \& technology  \textbf{11}(7),  12--12 (2022)

\bibitem{zhu2025eyebenchrigorousevaluationretinal}
Zhu, W., Dong, X., Li, X., Xiong, Y., Chen, X., Qiu, P., Vasa, V.K., Yang, Z., Su, Y., Dumitrascu, O., Wang, Y.: Eyebench: A call for more rigorous evaluation of retinal image enhancement (2025), \url{https://arxiv.org/abs/2502.14260}

\bibitem{zhu2024nnmobilenetrethinkingcnnretinopathy}
Zhu, W., Qiu, P., Chen, X., Li, X., Lepore, N., Dumitrascu, O.M., Wang, Y.: nnmobilenet: Rethinking cnn for retinopathy research (2024), \url{https://arxiv.org/abs/2306.01289}

\end{thebibliography}
\end{document}